\documentclass[journal]{IEEEtran}

\usepackage{color}
\usepackage{tabularray}
\usepackage{algorithm}
\usepackage{algpseudocode}
\usepackage{amsmath}
\usepackage{graphicx}

\newcommand{\etal}{\textit{et al}.}

\begin{document}
\title{WhisperNetV2: SlowFast Siamese Network For Lip-Based Biometrics}
\author{Abdollah Zakeri{$^{1}$},
Hamid Hassanpour{{$^{1}$}},
Mohammad Hossein Khosravi{{$^{2}$}},
Amir Masoud Nourollah{{$^{1}$}}
        
{$^{1}$}Faculty of Computer Engineering, Shahrood University of Technology  \\
{$^{2}$}Faculty of Electrical and Computer Engineering, University of Birjand  
}
\maketitle
\begin{abstract}
Lip-based biometric authentication (LBBA) has attracted many researchers during the last decade. The lip is specifically interesting for biometric researchers because it is a twin biometric with the potential to function both as a physiological and a behavioral trait. Although much valuable research was conducted on LBBA, none of them considered the different emotions of the client during the video acquisition step of LBBA, which can potentially affect the client’s facial expressions and speech tempo. We proposed a novel network structure called WhisperNetV2, which extends our previously proposed network called WhisperNet. Our proposed network leverages a deep Siamese structure with triplet loss having three identical SlowFast networks as embedding networks. The SlowFast network is an excellent candidate for our task since the fast pathway extracts motion-related features (behavioral lip movements) with a high frame rate and low channel capacity. The slow pathway extracts visual features (physiological lip appearance) with a low frame rate and high channel capacity. Using an open-set protocol, we trained our network using the CREMA-D dataset and acquired an Equal Error Rate (EER) of 0.005 on the test set. Considering that the acquired EER is less than most similar LBBA methods, our method can be considered as a state-of-the-art LBBA method. 

\end{abstract}
\IEEEpeerreviewmaketitle

\section{Introduction}
Biometric methods for person verification and identification have attracted much attention among researchers in the last decade, thanks to the numerous advantages they have compared to old authentication methods like passwords and Personal Identity Numbers (PINs). Unlike the old methods mentioned above, biometric traits cannot be forgotten or transferred to another person. Furthermore, a client’s biometric data cannot be stolen, and they are pretty hard to replicate. Considering the aforementioned merits, biometric methods provide people authentication systems with more security. Although some solutions like two-factor authentication have been proposed to improve the security of passwords, these methods are less resistant to spoofing attacks compared to biometric-based authentication systems.

Biometric methods are divided into two main categories, namely physiological and behavioral traits. The former is defined as measurable and unique aspects of the human body which can serve as an identity, like face \cite{zhao_towards_2022}, fingerprint \cite{jain_-line_1997}, palm print \cite{zhang_online_2003}, hand geometry \cite{sanchez-reillo_biometric_2000} and iris \cite{ma_personal_2003}, while the latter is defined as any behavioral pattern that is done in a unique manner by individuals and hence, can be used as a means of identification. Common methods using behavioral traits are gait recognition, signature recognition, speaker identification, handwriting 

identification, and keystroke dynamics.
Using lips as a biometric trait has attracted much attention in the last decade due to their potential to be employed both as a physiological and a behavioral trait for different tasks such as person identification and verification. Visual features like the color and geometry of the lips can be considered a physiological trait, and the dynamics of the lips while speaking is a behavioral trait. Although two people might utter the exact same words and convey the same concept, the dynamics of their lips will not be the same. As a result, the combination of lips’ visual features and the dynamics of the lips while uttering a phrase provides enough uniqueness for a secure biometric system. 
Numerous valuable research articles proposed using the audio modality along with the video of the lips, or even as a single trait to provide the discrimination power required for a biometric system. However, there are several downsides to incorporating the audio modality like the detrimental effect of noisy audio on the overall accuracy of the biometric system. Additionally, incorporation of audio modality decreases the applicability of the biometric system in places in which silence is a requirement like in a library. Furthermore, speech-impaired people cannot use these systems which is a major drawback.
Due to the abovementioned reasons, the audio modality reduces the biometric authentication system's applicability; therefore, we use a visual-only approach in our proposed method.
Lip-based biometric authentication (LBBA) method is a very straightforward and practical way of authentication since it does not need any special equipment for the video acquisition, which can be done using the front-facing camera of a regular smartphone. Furthermore, this method can be implemented in a very light and efficient way to be used on mobile devices without any significant processing power.
LBBA has attracted many researchers because of its merits and advantages over other biometric methods. Biometric traits necessitate uniqueness for each person, and lips are proven to provide this uniqueness \cite{bandyopadhyay_feature_2013, tsuchihashi_studies_1974}. 

Human lip functions both as a physiological and a behavioral trait, considering that every person might utter the same words uniquely \cite{mason_role_2002,chowdhury_lip_2022}. LBBA is more robust and secure than other physiological-only biometric systems. Incorporating lip movements as a behavioral trait makes spoofing attacks almost impossible since the attackers must model a person's talking habits, which is a highly complex task and requires an abundance of data. Furthermore, the video acquisition process for LBBA does not require any special sensors or instruments and can be effortlessly done using a video camera. This feature makes LBBA more applicable than other biometric methods requiring specific means of measurement. Additionally, LBBA can be combined with other biometric methods such as face to increase the security and robustness of the biometric system. Moreover, LBBA is more hygienic than other biometric methods like fingerprint which require the client to touch the sensor.  
Most visual biometric systems encounter the challenge of variations in clients' appearance over time. In a facial verification system, for instance, the growth of facial hair in men and the variation of makeup in women can sometimes increase the error rates of the biometric system. To solve this issue, the visual biometric systems must be trained using a large enough dataset containing a fair percentage of the possible variations in the data that the system might face in practical applications. Although some physiological biometric methods like fingerprint and iris recognition do not suffer from this issue, the LBBA method will face this challenge given that the clients may have several emotional states at the time of authentication, which will potentially affect their lip movements during an utterance. This change in the lip dynamics will decrease the similarity between the video acquired for enrollment and the video acquired for authentication. Consequently, this issue will increase the False Rejection Rate (FRR) of the biometric system. One possible solution for decreasing the FRR of the system would be to decrease the threshold so that the system is less strict. Nevertheless, there is a tradeoff between False Acceptance Rate (FAR) and FRR, and changing the threshold in favor of decreasing the FRR would increase the FAR. 
Another solution is to train the system so that it becomes inattentive to these minor changes and attends only to the static characteristics. In the domain of LBBA, the previous solution is translated as training the system so that it becomes invariant to all the emotions that the client might have during the authentication video acquisition. This method is only possible if we have the required data to train the biometric system.
In this paper, we extend our previous work, called WhisperNet \cite{zakeri_whispernet_2021}, by improving the proposed network architecture by substituting the embedding network with the SlowFast architecture which captures both physiological and behavioural features of lip videos, resulting in better performance and higher accuracy. 

The main contributions of this paper to the field are as follows:
(i).	Addressing the challenges that visual biometric systems face due to different emotions that the clients might have during authentication video/image acquisition and proposing a possible solution.
(ii).	Improving the network architecture proposed in our previous WhisperNet research in order to acquire both more efficient performance and higher accuracy.
(iii).	Introducing a state-of-the-art architecture for LBBA which can be adapted to other domains such as face verification/identification.
In the following sections of this paper, we go through the previous related works in Section 2, introduce the dataset and the pre-processing methods in Section 3, introduce the proposed network architecture in Section 4, and test results and experiments are presented in Section 5. The conclusion, followed by suggestions for future works, are presented in Section 6.

\section{Literature Review}
The first step of pre-processing for an LBBA system is to localize the lip region using image segmentation. Lip segmentation methods are primarily classified into contour-based \cite{chan_automatic_1999} and clustering-based techniques \cite{fu_robust_2016, wang_lip_2002}. An unsupervised segmentation method was proposed by Chan that used Gaussian Mixture Models (GMMs) for lip localization \cite{chan_automatic_1999}. A Fuzzy C-Means (FCM) algorithm was used by Fu et al. in \cite{fu_robust_2016} for lip segmentation. Furthermore, Lu and Liu \cite{lu_lip_2018} used the Localized Active Contour Model (LACM) and two initial contours for lip segmentation.
The task of feature extraction from lip images is the 
next step in an LBBA system. Earlier research articles used low-level hand-crafted features like color, shape \cite{luettin_speaker_1996}, texture, and geometry \cite{broun_automatic_2002} and mid-level feature extraction methods like Principal Component Analysis (PCA) and Linear Discriminant Analysis (LDA) to extract visual features from the lip area. Behavioral features contained temporal data and were extracted using models such as Hidden Markov Model (HMM) and GMM \cite{luettin_speaker_1996}. With the rise of deep learning techniques, Convolutional Neural Networks (CNNs) replaced the old-fashioned hand-crafted feature extraction techniques for visual feature extraction. The temporal data (lip movement among video frames) was modeled using RNNs like LSTM or GRU since the movements of the lips can be modeled as sequential data. In \cite{luettin_speaker_1996}, authors defined an Active Shape Model (ASM) based on lip boundary and intensity parameters. Furthermore, they used HMM and GMM for speaker identification and achieved 97.9\% accuracy on a small dataset containing only 12 clients. Wark et al. \cite{wark_use_2000} used lip contour profiles and a multi-stream HMM for identification and acquired 80\% accuracy on the XM2VTS dataset. Lai et al. \cite{lai_visual_2016} proposed using Sparse Coding (SC) to characterize the movements in the lip region followed by a max pooling on a spatio-temporal hierarchical structure in order to produce the final features. They tested their method on a private dataset of 40 clients and achieved a Half Total Error Rate (HTER) of 0.46\%. Wright and Stewart \cite{wright_one-shot-learning_2019} trained a Siamese network using the LipNet structure containing STCNN and biGRU layers as the embedding network for the task of speaker verification under a closed-set Lausanne protocol (all subjects are enrolled during training) and achieved 1.03\% EER on the XM2VTS dataset. Their next article \cite{wright_understanding_2020} tested their architecture under a more restrictive open-set protocol (new subjects are enrolled during validation and test) and achieved an EER of 1.65\%. Kuang \etal \cite{10163930} introduced LipAuth, leveraging unique spatial-temporal features of human lips, achieving a 99.24\% accuracy for user authentication and demonstrating robustness against video replay and mimic attacks. Similarly, Koch and Grbić \cite{KOCH2024104900} addressed vulnerabilities in existing LBBA methods by introducing the GRID-CCP dataset and training a siamese neural network with 3D convolutions and recurrent neural network layers, achieving a False Acceptance Rate (FAR) of 3.2\% and a False Rejection Rate (FRR) of 3.8\%. An overview of existing LBBA methods and datasets are provided in tables \ref{table1} and \ref{table2} accordingly.

\definecolor{Black}{rgb}{0,0,0}
\begin{table*}
\caption{An overview of the existing LBBA methods}
\label{table1}
\centering
\begin{tblr}{
  width = \linewidth,
  colspec = {Q[56]Q[144]Q[394]Q[129]Q[83]Q[127]},
  cell{6}{1} = {r=2}{},
  cell{6}{2} = {r=2}{},
  cell{6}{3} = {r=2}{},
  hline{1,13} = {-}{},
  hline{2-6,8-12} = {-}{Black},
  hline{7} = {4-6}{Black},
}
\textbf{Year} & \textbf{Proposed By}                   & \textbf{Method}                                                 & \textbf{Dataset}      & \textbf{\#Clients} & \textbf{Performance} \\
\textbf{1996} & Luettin \etal  \cite{luettin_speaker_1996}        & Active Shape Model (ASM) + HMM  GMM                             & Private               & 12                 & 97.9\% Acc           \\
\textbf{2000} & Wark \etal  \cite{wark_use_2000}            & Lip contour profiles + multi-stream HMM                         & M2VTS \cite{pigeon_m2vts_1997}            & 37                 & 80\% Acc             \\
\textbf{2002} & Broun \etal  \cite{broun_automatic_2002}           & Geometric features + polynomial-based model                     & XM2VTS \cite{messer_xm2vtsdb_2000}           & 295                & 6.3\% HTER           \\
\textbf{2012} & Chan \etal  \cite{chan_local_2012}            & LOCP - TOP                                                      & XM2VTS                & 295                & 0.36\% HTER          \\
\textbf{2013} & {Bakry and\\Elgammal \cite{bakry_mkpls_2013}}             & Nonlinear mapping + Kernel Partial Least Squares                & AVLetters \cite{matthews_extraction_2002}        & 10                 & 42.82\% Acc          \\
              &                                        &                                                                 & OuluVS \cite{anina_ouluvs2_2015}           & 52                 & 62.34\% Acc          \\
\textbf{2014} & {Liu and\\Cheung \cite{liu_learning_2014}}                 & Multi-boosted HMMs                                              & Private               & 46                 & 3.91\% EER           \\
\textbf{2016} & Lai \etal  \cite{lai_visual_2016}             & {Joint spatio-temporal~sparse\\coding~and hierarchical pooling} & Private               & 40                 & 0.46\% HTER          \\
\textbf{2019} & {Wright and\\Stewart \cite{wright_one-shot-learning_2019}}             & Siamese(STCNN+BiGRU)                                            & XM2VTS                & 295                & 1.03\% EER           \\
\textbf{2020} & {Wright and\\Stewart \cite{wright_understanding_2020}}             & Siamese(STCNN+BiGRU)                                            & XM2VTS                & 295                & 1.65\% EER           \\
\textbf{2021} & {Zakeri and\\Hassanpour \cite{zakeri_whispernet_2021}\textbf{}} & Siamese(STCNN+BiGRU)\textbf{}                                   & CREMA-D \cite{cao_crema-d_2014}\textbf{} & 88\textbf{}        & 95.41\% Acc \\
\textbf{2024} & {Koch and Grbic \cite{KOCH2024104900}} & Siamese(3D CNN+RNN) & GRID-CCP & - & 3.2\% FAR, 3.8\% FRR \\ \hline
\textbf{2024} & \text{Kuang \etal \cite{10163930}} & LipAuth & \text{Custom Dataset} & 50 & 99.24\% \text{ Acc} \\
\hline

\end{tblr}
\end{table*}

\begin{table*}
\centering
\caption{An overview of the existing LBBA datasets}
\label{table2}
\begin{tblr}{
  width = \linewidth,
  colspec = {Q[77]Q[265]Q[300]Q[163]Q[125]},
  hlines,
  hline{2-9} = {-}{Black},
}
\textbf{Year} & \textbf{Proposed By}             & \textbf{Name}        & \textbf{Resolution} & \textbf{\#Clients} \\
\textbf{1999} & Messer \etal \cite{messer_xm2vtsdb_2000}     & XM2VTS               & 720×576             & 295                \\
\textbf{2002} & Patterson \etal  \cite{patterson_cuave_2002} & CUAVE                & 64×32               & 36                 \\
\textbf{2005} & Fox \etal  \cite{fox_valid_2005}       & VALID                & 720×576             & 106                \\
\textbf{2005} & Peer \etal \cite{peer_computer_2005}                        & CVL Face Database    & 640×480             & 114                \\
\textbf{2015} & Anina \etal  \cite{anina_ouluvs2_2015}     & OuluVS2              & 44×30               & 52                 \\
\textbf{2016} & Bakshi \etal \cite{bakshi_nitrlipv1_2016}      & Nitrlipv1            & 180×180             & 109                \\
\textbf{2017} & Raman \etal  \cite{raman_acquisition_2017}     & NITRLipV2 (MobioLip) & 72×72               & 20                 
\end{tblr}
\end{table*}

\begin{algorithm*}
\caption{Lip Landmarks Extraction and Bounding Box Adjustment}
\label{code1}
\begin{algorithmic}[1]
\For{video in Dataset}
    \For{frame in video}
        \State $Landmarks \gets \{(x, y)_i\}_{i=1,...,24}$ \Comment{Extract frame's lip landmarks using FaceRecognition Library}
        \State $x_{\min} \gets \min\{x \mid (x, y)_i \in Landmarks\}$
        \State $x_{\max} \gets \max\{x \mid (x, y)_i \in Landmarks\}$
        \State $y_{\min} \gets \min\{y \mid (x, y)_i \in Landmarks\}$
        \State $y_{\max} \gets \max\{y \mid (x, y)_i \in Landmarks\}$
        \State $BoundingBoxTopLeft \gets (x_{\min}, y_{\min})$
        \State $BoundingBoxBottomRight \gets (x_{\max}, y_{\max})$
        \State $BoundingBoxWidth \gets x_{\max} - x_{\min}$
        \State $BoundingBoxHeight \gets y_{\max} - y_{\min}$
        \State $BoundingBoxAspectRatio \gets \frac{BoundingBoxWidth}{BoundingBoxHeight}$
        \If{$BoundingBoxAspectRatio < AR$} \Comment{AR is the aspect ratio with the highest absolute frequency, which is 5/3 in our case.}
            \State $BoundingBoxWidth \gets BoundingBoxHeight \times AR$
        \Else
            \State $BoundingBoxHeight \gets BoundingBoxWidth \times \left(\frac{1}{AR}\right)$
        \EndIf
        \State $NewFrame \gets$ Crop the current frame according to the calculated bounding box and then resize to 30x18
    \EndFor
\EndFor
\end{algorithmic}
\end{algorithm*}
\begin{figure*}[ht]
    \centering
    \includegraphics[width=\linewidth]{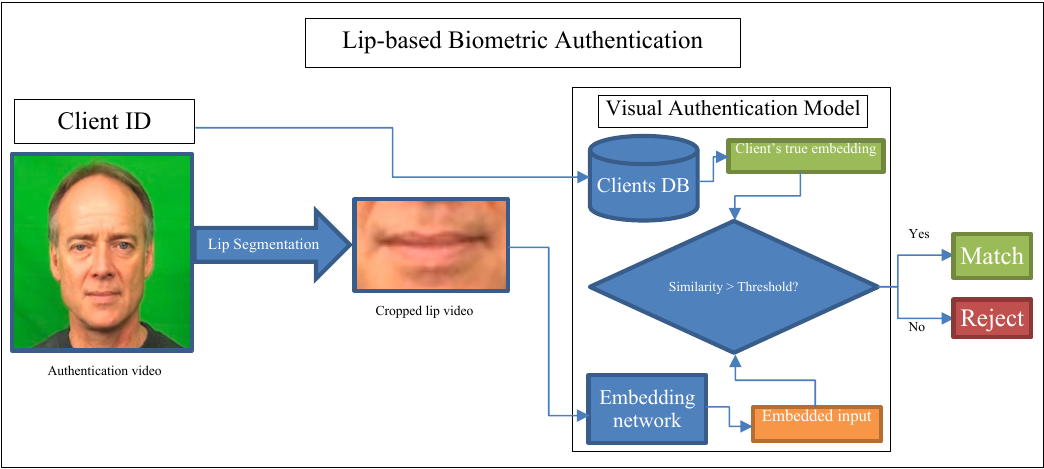}
    \caption{Flowchart of our proposed method for pre-processing and visual speaker authentication}
    \label{fig2}
\end{figure*}

\section{Dataset and pre-processing}
Numerous datasets are available that can be used to train an LBBA network \cite{messer_xm2vtsdb_2000, anina_ouluvs2_2015, patterson_cuave_2002, fox_valid_2005, peer_computer_2005, bakshi_nitrlipv1_2016,raman_acquisition_2017}. To the best of the authors’ knowledge, none of them include emotion-related data concerning the subjects’ facial expressions or variations in utterance and speech tempo due to different emotional states. Hence, we used the CERMA-D dataset \cite{cao_crema-d_2014} to train our network, which contains 7,442 videos of 91 subjects uttering 12 phrases in 6 different emotional states, including neutral, happy, sad, anger, disgust, and fear. Nevertheless, there were data shortages for three clients, so we excluded them from the dataset, and therefore, we were left with 6,336 videos and 88 clients. Although the size of our dataset is smaller than other datasets like XM2VTS \cite{messer_xm2vtsdb_2000}, our dataset contains emotional data that provides various utterances of the same phrase by the same subject but with different facial expressions and speech tempos. Considering that the network is trained assuming that the different utterances of the same phrase by the same person are equivalent, the network gradually becomes invariant to the small visual changes that are caused by different emotions. An overview of the datasets applicable in LBBA is listed in table \ref{table2}.

Out of the 88 people in our dataset, 66 were assigned to the training set, 11 to the validation set, and 11 to the test set. Since the clients in the test set are not enrolled during training, we used an open-set protocol \cite{cao_crema-d_2014}. 
Since the dataset included full-face videos, we needed to crop each frame of every video to the lip region so that the network would not take advantage of any data other than lips. We used an open-source Python library called Face Recognition \cite{geitgey_machine_2020} to perform this task. We extracted the lip landmarks from every frame of each video, and then using these landmarks, we selected a bounding box containing only the lips and cropped the whole frame into that bounding box. After cropping, we resized each cropped image to the size of the smallest lip image (30×18). 

Furthermore, the bounding boxes were all chosen in a way that they would have the same aspect ratio in order to minimize the interpolations during resizing. The complete algorithm for pre-processing of the database is presented in Algorithm \ref{code1}.

\section{Proposed Method}

There are two common steps in biometric authentication systems. The first step is called enrollment and includes generating the corresponding embedding for each client based on their biometric features and saving this embedding along with a unique ID in the clients’ database. The next step is authentication, in which the system obtains biometric information from the client and generates the corresponding embedding based on this obtained information. Next, the system compares this newly generated embedding with the one stored in the database, and if the two embeddings’ similarity surpasses a static threshold, the person is successfully authenticated. Aside from the pre-processing and data preparations, which are specific to our method, LBBA has the same two steps. A flowchart of our proposed LBBA method is demonstrated in Fig. \ref{fig2}.

\begin{figure}[ht]
    \centering
    \includegraphics[width=\linewidth]{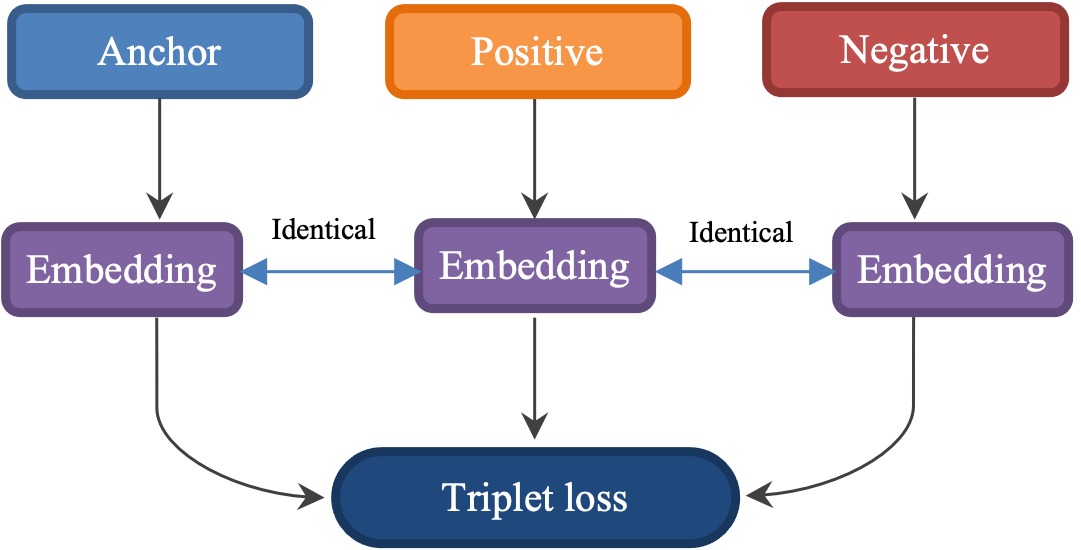}
    \caption{Structure of a Siamese Network}
    \label{fig3}
\end{figure}

\begin{figure*}[ht]
    \centering
    \includegraphics[width=\linewidth]{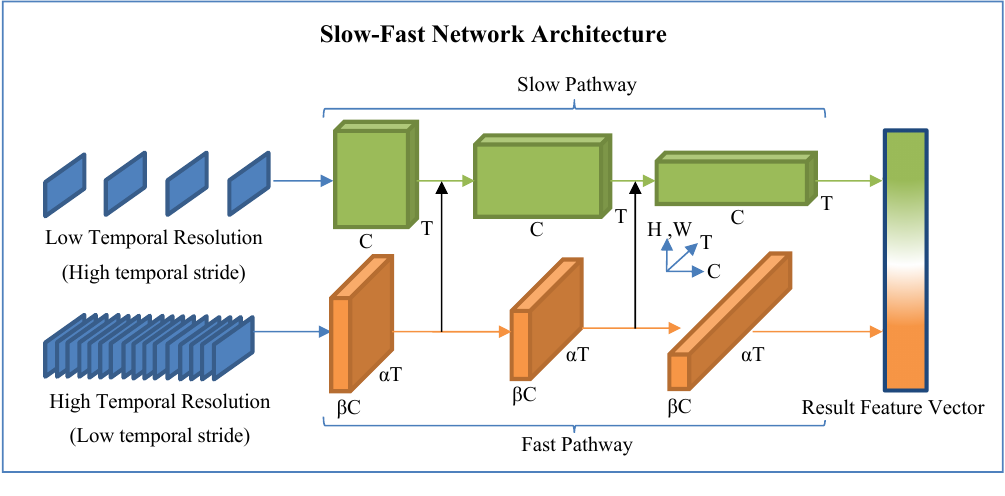}
    \caption{Slow-Fast Network Architecture}
    \label{fig4}
\end{figure*}

The embedding is a network mapping the input data to latent space, and its output is called the embedding or the encoding. In order to train this network, we used a Siamese architecture with the triplet loss function \cite{bromley_signature_1993}, \cite{chopra_learning_2005}. The Siamese architecture consists of three identical branches that take triplets of input videos and generate the corresponding embeddings for each item of the triplet tuple. The structure of the Siamese network is presented in Fig. \ref{fig3}.  Each triplet includes three videos, namely the anchor (A), the positive (P), and the negative (N). The anchor and positive are two videos of the same person uttering the same phrase. The negative can be any video that does not fit into the definition of positive. The corresponding embeddings for A, P, and N are generated using three identical embedding networks. Using the triplet loss function, the network gradually learns to generate embeddings so that the distance between A and P embeddings is less than the distance between A and N embeddings. The triplet loss is defined as follows:

\begin{equation}
\mathcal{L}(A, P, N) = \max(D(A, P) - D(A, N) + \alpha, 0)
\end{equation}

The loss function ensures a minimum margin with the value of $\alpha$ between any of the anchor-positive and the anchor-negative distances. The distance function D can be any distance measure like cosine or Euclidian distance. We chose cosine similarity as the distance function and the value of 0.7 for the margin parameter. The cost function is calculated as follows:

\begin{equation}
\mathcal{S} = \sum_{i=1}^{m} \mathcal{L}(A^{(i)}, P^{(i)}, N^{(i)}) 
\end{equation}
\quad (\text{$m$ denotes the total number of samples in the dataset})\\

\begin{figure*}[ht]
    \centering
    \includegraphics[width=\linewidth]{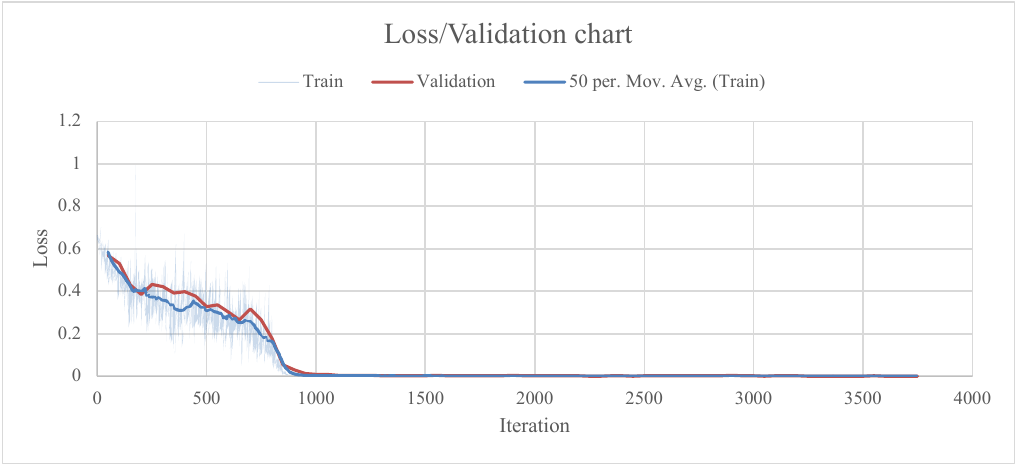}
    \caption{Training loss curve for our proposed network. Due to fluctuations in the loss values, the moving average for loss values is plotted.}
    \label{fig5}
\end{figure*}

We used SlowFast architecture (Fig. \ref{fig4}) as the embedding network \cite{feichtenhofer_slowfast_2019}. This network was initially proposed for video recognition tasks. It passes the same video into two processing pathways: the slow pathway and the fast pathway. The fast pathway processes the input video with a high Frame Per Second (FPS) (low temporal stride) to extract motion-related data from consecutive frames. The slow pathway processes the input video with low FPS (high temporal stride) to extract static visual features. Although the fast pathway has a higher temporal resolution, it has lower channel capacity than the slow pathway.
There are two main hyperparameters for SlowFast networks, namely $\alpha$ and $\beta$. The $\alpha$ parameter represents the temporal stride ratio between the slow and fast pathways. For example, if the temporal strides of the fast and slow pathways are 2 and 16 accordingly, then $\alpha$=8, which is the typical value for this parameter according to the literature \cite{feichtenhofer_slowfast_2019} The $\beta$ parameter represents the channel ratio between the slow and fast pathways ($\beta$<1). To sum it up, both the slow and fast pathways are identical CNNs, but the fast pathway has a low temporal stride (high temporal resolution) and low channel capacity. In contrast, the slow pathway has a high temporal stride (low temporal resolution) and high channel capacity.
As mentioned before, the lip can be used as a twin biometric, considering its physiological and behavioral features. This capability of the lips makes the SlowFast networks an ideal candidate for the task of LBBA since, on the one hand, the fast pathway extracts motion-related features from the lip video, which are the behavioral traits related to lip movements during an utterance. These motion-related data do not need high channel capacity, making the fast pathway an excellent candidate for behavioral feature extraction. On the other hand, the physiological features are static visual features that do not change much during the video. Hence, the extraction of these features does not need high temporal resolution, but its high channel capacity would be beneficial, and this is the exact description of the slow pathway. At the final layer of the SlowFast architecture, the outputs of the two pathways are concatenated, forming our final embedding. This SlowFast embedding network uses physiological and behavioral lip features to generate the final embedding vector.
Additionally, several lateral connections between the two pathways fuse the information of the two pathways before the final concatenation layer, which makes each pathway aware of the representations learned by the other. However, the authors of \cite{feichtenhofer_slowfast_2019} found similar results while using bidirectional and unidirectional lateral connections, and therefore, in the implementation of this network, these lateral connections are unidirectional connections from the fast pathway to the slow pathway. The result of each “stage” of the fast pathway is concatenated with the result of the previous stage before being entered to the next stage of the slow pathway. According to \cite{feichtenhofer_slowfast_2019}, since the two pathways have different temporal dimensions, the lateral connections must perform a transformation to match them.

One of the crucial tasks in training Siamese networks with triplet loss is the selection of triplets. Wright and Stewart \cite{wright_understanding_2020} classified the triplets into three categories:
\begin{itemize}
    \item[(i)] Easy triplets are triplets in which the distance between the anchor and positive embeddings is less than the distance between anchor and negative:
    \begin{equation}
    D(A,P) + \alpha < D(A,N)
    \end{equation}

    \item[(ii)] Semi-hard triplets are triplets in which the positive is closer to the anchor than the negative but still generate a positive loss since the negative stays within the margin:
    \begin{equation}
    D(A,P) < D(A,N) < D(A,P) + \alpha
    \end{equation}

    \item[(iii)] Hard triplets are triplets in which the distance between the negative and anchor is less than the distance between the positive and the anchor:
    \begin{equation}
    D(A,P) \geq D(A,N)
    \end{equation}
\end{itemize}

Using only hard triplets stops the network from converging, and using only easy triplets will not contribute much to decreasing the loss value. So we selected triplets randomly from all the possible train triplets:

\begin{equation}
\begin{aligned}
\underbrace{(66 \times 12 \times 6)}_{\text{Anchor}} & \times \underbrace{5}_{\text{Positive}} \times \\
\underbrace{\left(\underbrace{(65 \times 12 \times 6)}_{\substack{\text{Other People,} \\ \text{all Phrases}}} + \underbrace{(1 \times 11 \times 6)}_{\substack{\text{Same Person,} \\ \text{Other Phrases}}}\right)}_{\text{Negative}} &= 112,764,960
\end{aligned}
\end{equation}

Since the client should not be authenticated if they do not utter the same phrase in the anchor and the positive videos, all other utterances of the remaining phrases done by the same client are considered negative.

There are two challenging sets of triplets that are highly beneficial for the embedding network. The first are triplets in which the negative video belongs to the same client as the positive and the anchor, but the uttered phrase is different. While training on these triplets, the slow pathways of the embedding networks produce almost the same output for all three inputs. As a result, the network should only lean on the output of the fast pathway (behavioral features) to decrease the loss value, which improves the fast pathway.

The other set of challenging triplets are triplets in which the negative video has the same phrase as the anchor and the positive but uttered by other clients. In contrast with the former set of triplets, the fast pathways of embedding networks will generate similar embeddings. The network must use only the slow pathway (physiological features) to decrease the triplet loss value since the difference between physiological features of anchor and negative is more than the difference between their behavioral features. Consequently, the slow pathway of the embedding network will be improved by using these triplets.

\section{Experiments and Results}
All training and evaluations were performed on a linux-based system with 2×Nvidia A100 GPU’s, 512 GB DDR4 RAM, and an AMD EPYC 7H12 64-core processor. With the aforementioned configuration, a single inference step including I/O tasks, preprocessing, and feeding the input videos to the embedding network to get the final result, takes $\approx$2.37 seconds (1.43s for reading pair of videos, 0.42s for preprocessing, 0.52 for generating embeddings and comparison).

We trained our network for ~3800 iterations, however the criterion for ending training (I<Threshold) was met before 2000 iterations. In each iteration, we sampled a random batch (batch size = 512) from the training dataset and due to a large number of training samples, there was no overfitting during the training session. Training/Validation loss values are plotted in Fig. \ref{fig5}.

There are three standard metrics for the evaluation of biometric verification systems:

\begin{enumerate}
    \item \textbf{False Rejection Rate (FRR):} Percentage of the times a biometric verification system denies access to an authorized client. This error is known as Type I error in statistical terms.
    
    \begin{equation}
        FRR = \frac{\# \text{ of denied authentic requests}}{\# \text{ of all requests}}
    \end{equation}
    
    \item \textbf{False Acceptance Rate (FAR):} Percentage of the times a biometric verification system grants access to an imposter. This error is known as Type II error in statistical terms.
    
    \begin{equation}
        FAR = \frac{\# \text{ of granted imposter requests}}{\# \text{ of all requests}}
    \end{equation}

    \item \textbf{Crossover Error Rate (CER):} Point where the FAR and FRR are equal. This metric describes the overall performance of the biometric verification system.

\end{enumerate}

\begin{figure}[ht]
    \centering
    \includegraphics[width=\linewidth]{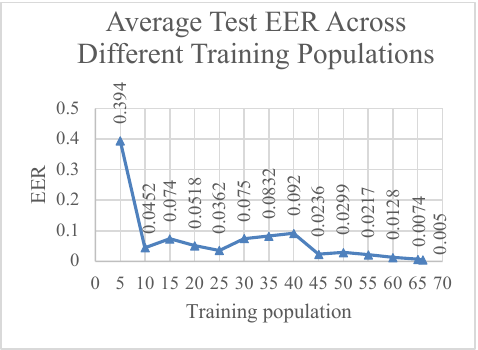}
    \caption{Training loss curve for our proposed network. Due to fluctuations in the loss values, the moving average for loss values is plotted.}
    \label{fig6}
\end{figure}

\begin{figure}[ht]
    \centering
    \includegraphics[width=\linewidth]{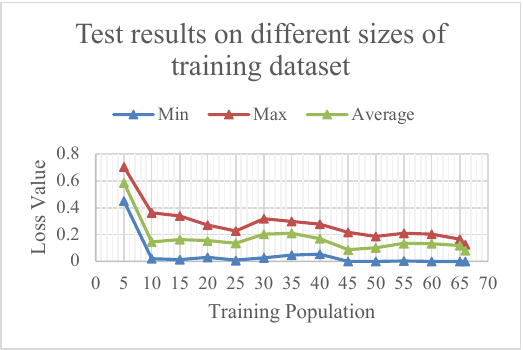}
    \caption{Training loss curve for our proposed network. Due to fluctuations in the loss values, the moving average for loss values is plotted.}
    \label{fig7}
\end{figure}

As we witnessed in our experiments, there is a tradeoff between FAR and FRR while setting the threshold value for the biometric system. In a perfect biometric verification system, both FAR and FRR would be 0\% for all of the possible threshold values, and as a result, the EER would equal 0\%.
To test our proposed method, we selected 11 unseen clients and generated the corresponding triplets for them. According to a formula similar to (6), the number of test triplets would be 3,112,560, including hard, semi-hard, and simple triplets. These triplets were sampled into batches of size 512, and for each triplet, the cosine similarities for anchor positive and anchor negative pairs were calculated. For every value of similarity threshold between 0 and 1, with a precision of 0.001, we calculated the FAR and FRR values. At the threshold of 0.983, the FAR and FRR values were equal and had the value of 0.005, which would be the EER of our proposed method. 
Additionally, to assess the performance of the network with different training populations, we sampled random subsets of training people and generated the training triplets accordingly. For comparison purposes, all trained networks were evaluated using a fixed test set (size=11 people). The results suggested that the overall performance of the network improves with increasing the size of the training set. For each training population size, the network was trained for 20 sessions and the training and validation datasets were shuffled before each session. The Average EER values for each training population are demonstrated in Fig. \ref{fig6}, and the loss values for the same experiments are plotted in Fig. \ref{fig7}. 

\begin{figure}[t]
    \centering
    \includegraphics[width=\linewidth]{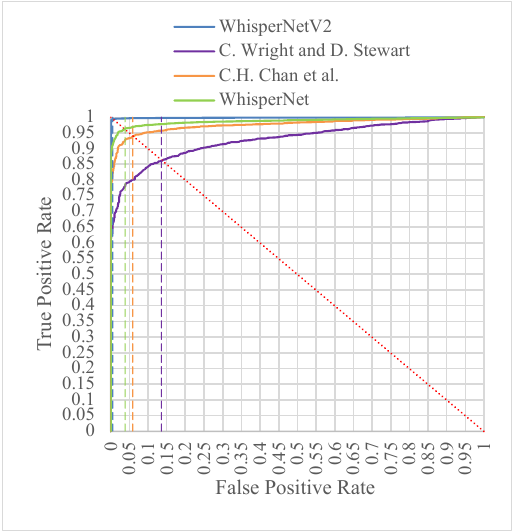}
    \caption{Training loss curve for our proposed network. Due to fluctuations in the loss values, the moving average for loss values is plotted.}
    \label{fig8}
\end{figure}

We evaluated the performance of our newly proposed method alongside two established approaches (Wright and Stewart \cite{wright_understanding_2020}, and Chan et al. \cite{chan_local_2012}), as well as our previous model, on the CREMA-D dataset. All methods were subjected to the same training criteria, parameters, and pre-processing steps, with only differences in input video sizes to match each method's requirements.
The results demonstrated in Fig. \ref{fig8} clearly indicate that our proposed method excels in terms of accuracy, with an exceptionally low Equal Error Rate (EER) of 0.005. In contrast, Wright and Stewart \cite{wright_understanding_2020} and Chan et al. \cite{chan_local_2012} yielded higher EER values of 0.136 and 0.059, respectively. Our previous model, represented by an EER of 0.038 in the ROC curve, performs better than the other two methods.
One crucial aspect to consider is the complexity of the CREMA-D dataset, which features utterances of the same phrase expressed with varying emotions. This variability introduces complexity in facial expressions and speech tempo, which demands a certain level of model sophistication to effectively capture. Models lacking this complexity tend to be highly biased and, as evident in the results, exhibit degraded performance.
In essence, the superior performance of our proposed method highlights the importance of finding the right balance between security and usability. Achieving low EER values signifies a robust security stance, while simultaneously ensuring that the model can handle the dataset's inherent complexity enhances usability. Our model strikes this balance effectively, outperforming both previous iterations and competing methods in this challenging biometric authentication task on the CREMA-D dataset. This underscores the significance of comprehensive model design when considering real-world applications that involve complex data.

 In the context of our research findings, it's crucial to examine the concepts of true positives (TP) and false positives (FP) as they relate to the results. True positives represent instances where our biometric authentication system correctly identified a legitimate user, thus providing a measure of the system's accuracy and security. Conversely, false positives signify cases where the system erroneously accepted an imposter as a genuine user, which can lead to potential security vulnerabilities and usability issues. Comparing these metrics with those from previous studies, our proposed method demonstrates a remarkable advantage. The lower false positive rate in our model, evident in the considerably lower Equal Error Rate (EER) of 0.005, underscores its heightened security, as it effectively minimizes the risk of unauthorized access. In contrast, previous methods, such as Wright and Stewart's \cite{wright_understanding_2020} with an EER of 0.136 and Chan et al.'s \cite{chan_local_2012} with an EER of 0.059, exhibited higher false positive rates, indicating a higher susceptibility to imposter acceptance. These findings emphasize the significance of our model's enhanced security and its potential for real-world applications where minimizing false positives is paramount.
In our research paper, our latest model demonstrates notable improvements over its predecessor, achieving a reduction of 0.033 in the Equal Error Rate (EER) while simultaneously enhancing computational efficiency. The architecture of our previous embedding network consisted of Spatio-Temporal Convolutional Neural Networks (STCNNs) followed by Gated Recurrent Units (GRUs), totaling 76,749,124 trainable parameters. In contrast, our current proposed embedding network employs the more streamlined SlowFast network with 52,314,144 trainable parameters, resulting in accelerated training and inference processes.
Moreover, our previous method relied on a list of 24 lip landmark points to generate embeddings, making the network susceptible to errors. These errors in lip landmark extraction had a cascading effect on the embedding network, compromising both the overall performance and robustness of the model. To address this issue, our present study eliminates the lip landmark extraction step, effectively resolving this source of error.

\section{Conclusion and Future Works}

This study presents a deep Siamese network for LBBA using the SlowFast network as the embedding network. The SlowFast architecture has a fast pathway with low temporal stride and low channel capacity, which extracts behavioral features, and a slow pathway with high temporal stride and high channel capacity extracting physiological features from the lip authentication video. We acquired an EER of 0.005 on the test set containing 11 unseen clients. Our model improves upon the performance, robustness, and efficiency of its predecessor by decreasing both EER and number of trainable parameters. Although the propoed method outperforms similar LBBA methods and the acquired results seems promising for real-world applications, some challenges have not been mentioned in this paper, like the performance of the proposed method under various lighting conditions and different resolutions. Another challenge in real-world applications would be the capacity of the network. Several factors directly impact the capacity of our proposed network, namely the length of the uttered phrase in enrollment and authentication videos and the size of the embedding vector, which is the output of our SlowFast embedding network. Our current dataset has certain limitations, and performing evaluations on larger publicly available datasets would make our method more reliable and robust. We plan to create a more extensive dataset for the LBBA task in the near future to facilitate research in this field and provide a benchmark for the fair comparison of different LBBA methods.

\bibliographystyle{IEEEtran}
\bibliography{main}

\end{document}